\documentclass{article}

\usepackage{docmute}

\usepackage{arxiv}

\usepackage{graphicx}

\usepackage[utf8]{inputenc}
\usepackage[T1]{fontenc}

\usepackage{amsmath}
\usepackage{amsfonts}
\usepackage{bm}
\usepackage{comment}
\usepackage{fancybox}
\usepackage{framed}
\usepackage{color}
\usepackage{multicol}
\usepackage{multirow}
\usepackage{hyperref}
\usepackage{url}

\usepackage{enumitem}

\usepackage{physics}

\usepackage[whole,substmingoth]{bxcjkjatype}

\usepackage{algorithm}
\usepackage[noend]{algpseudocode}
\usepackage{algorithmicx}

\usepackage{tabularx}
\newcolumntype{Y}{>{\centering\arraybackslash}X}

\usepackage{subcaption}

\usepackage{siunitx}

\usepackage[dvipsnames]{xcolor}

\expandafter\def\expandafter\UrlBreaks\expandafter{\UrlBreaks
  \do\a\do\b\do\c\do\d\do\e\do\f\do\g\do\h\do\i\do\j%
  \do\k\do\l\do\m\do\n\do\o\do\p\do\q\do\r\do\s\do\t%
  \do\u\do\v\do\w\do\x\do\y\do\z\do\A\do\B\do\C\do\D%
  \do\E\do\F\do\G\do\H\do\I\do\J\do\K\do\L\do\M\do\N%
  \do\O\do\P\do\Q\do\R\do\S\do\T\do\U\do\V\do\W\do\X%
  \do\Y\do\Z}

\DeclareMathOperator{\sgn}{sgn}

\algrenewcommand\algorithmicindent{1.0em}

\algnewcommand\algorithmicforeach{\textbf{for each}}
\algdef{S}[FOR]{ForEach}[1]{\algorithmicforeach\ #1\ \algorithmicdo}

\algnewcommand\AlgAnd{\textbf{and} }
\algnewcommand\AlgOr{\textbf{or} }
\algnewcommand\AlgContinue{\textbf{Continue}}
\algnewcommand\AlgBreak{\textbf{break}}

\algrenewcommand\textproc{}

\algnewcommand{\Initialize}[1]{
	\State \textbf{Initialize:}
 	\State \hspace*{\algorithmicindent}\parbox[t]{0.8\linewidth}{\raggedright #1}}

\algnewcommand{\LeftComment}[1]{
  \Statex $\triangleright$ #1 \hfill}

\algnewcommand{\IIf}[1]{\State\algorithmicif\ #1\ \algorithmicthen}
\algnewcommand{\EndIIf}{\unskip}

\def\BibTeX{{\rm B\kern-.05em{\sc i\kern-.025em b}\kern-.08em
  T\kern-.1667em\lower.7ex\hbox{E}\kern-.125emX}}

\newcommand{\MethodName}{ElasticZO}
\newcommand{\MethodNameInt}{ElasticZO-INT8}

\title{{\MethodName}: A Memory-Efficient On-Device Learning with Combined Zeroth- and First-Order Optimization}

\renewcommand{\shorttitle}{{\MethodName}: Memory-Efficient On-Device Learning}

\author{%
  \href{https://orcid.org/0000-0001-8534-2381}%
  {\includegraphics[scale=0.06]{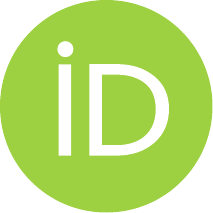}%
  \hspace{1mm}Keisuke Sugiura}\\
  Keio University\\
  3-14-1 Hiyoshi, Kohoku-ku, Yokohama, Japan\\
  \texttt{sugiura@arc.ics.keio.ac.jp}\\
  \And
  \href{https://orcid.org/0000-0001-9578-3842}%
  {\includegraphics[scale=0.06]{orcid.pdf}%
  \hspace{1mm}Hiroki Matsutani}\\
  Keio University\\
  3-14-1 Hiyoshi, Kohoku-ku, Yokohama, Japan\\
  \texttt{matutani@arc.ics.keio.ac.jp}
}

\hypersetup{
  pdftitle = {{\MethodName}: A Memory-Efficient On-Device Learning with Combined Zeroth- and First-Order Optimization},
  pdfsubject = {cs.LG},
  pdfauthor = {Keisuke Sugiura, Hiroki Matsutani},
  pdfkeywords = {On-Device Learning, Integer-Only Training, Deep Learning, Zeroth-Order Optimization}
}

\begin{document}

\maketitle



\begin{abstract}
Zeroth-order (ZO) optimization is being recognized as a simple yet powerful alternative to standard backpropagation (BP)-based training.
Notably, ZO optimization allows for training with only forward passes and (almost) the same memory as inference, making it well-suited for edge devices with limited computing and memory resources.
In this paper, we propose ZO-based on-device learning (ODL) methods for full-precision and 8-bit quantized deep neural networks (DNNs), namely {\MethodName} and {\MethodNameInt}.
{\MethodName} lies in the middle between pure ZO- and pure BP-based approaches, and is based on the idea to employ BP for the last few layers and ZO for the remaining layers.
{\MethodNameInt} achieves integer arithmetic-only ZO-based training for the first time, by incorporating a novel method for computing quantized ZO gradients from integer cross-entropy loss values.
Experimental results on the classification datasets show that {\MethodName} effectively addresses the slow convergence of vanilla ZO and shrinks the accuracy gap to BP-based training.
Compared to vanilla ZO, {\MethodName} achieves 5.2--9.5\% higher accuracy with only 0.072--1.7\% memory overhead, and can handle fine-tuning tasks as well as full training.
{\MethodNameInt} further reduces the memory usage and training time by 1.46--1.60x and 1.38--1.42x without compromising the accuracy.
These results demonstrate a better tradeoff between accuracy and training cost compared to pure ZO- and BP-based approaches, and also highlight the potential of ZO optimization in on-device learning.
\end{abstract}




\section{Introduction} \label{sec:intro}
First-order (FO) optimization algorithms with backpropagation (BP)~\cite{DavidERumelhart86,YannLeCun89A,YannLeCun89B,JohnDuchi11,DiederikPKingma15} have been predominantly used for training deep neural networks (DNNs) thanks to the wide support in popular DL frameworks.
While BP provides a systematic way to compute FO gradients via chain-rule by traversing the computational graph, it needs to save intermediate activations as well as gradients (with respect to parameters), which incurs considerably higher memory requirements than inference~\cite{SadhikaMalladi23} and may pose challenges for deployment on the memory-constrained platforms (e.g., Raspberry Pi Zero).
Besides, advanced FO optimizers consume extra memory to store optimizer states such as momentum (running average of past gradients) and a copy of the trainable parameters.
Given this situation, in the recent literature, zeroth-order (ZO) optimization has seen a resurgence of interest as a simple yet powerful alternative to FO methods~\cite{SijiaLiu20,YihuaZhang24}.

One notable feature of ZO methods is that it only requires two forward passes per input during training.
Since ZO gradients can be obtained from DNN outputs (loss values), ZO-based approach becomes an attractive choice when FO gradients are infeasible to obtain or not available (e.g., non-differentiable loss functions).
It has been applied to a wide range of practical applications including black-box adversarial attacks~\cite{PinYuChen17,HaishanYe18,ChunChenTu19} (where attackers only have an access to DNN inputs and outputs), black-box defense~\cite{YimengZhang22,AochuanChen24}, neural architecture search~\cite{XiaoxingWang22,LunchenXie23}, sensor selection in wireless networks~\cite{SijiaLiu18A}, coverage maximization in cellular networks~\cite{PengchengHe22,PengchengHe24}, and reinforcement learning from human feedback~\cite{ZhiweiTang24,QiningZhang24}.
Since ZO methods bypass BP, they do not need to retain computational graphs as well as intermediate activations and gradients.
With a simple random seed trick proposed in \cite{SadhikaMalladi23}, ZO methods can train DNNs with nearly the same memory usage as inference and still achieve comparable accuracy to BP training, leading to a recent success in the fine-tuning of large-scale DNNs (e.g., LLMs)~\cite{SadhikaMalladi23,TanmayGautam24,ShuoranJiang24,YongLiu24,WentaoGuo24}.
ZO methods can be easily implemented by modifying a small part of the inference code, and can utilize the existing highly-optimized inference engines to boost training speed, which reduces the development cost and enables a rapid deployment.
Despite these advantages over BP, ZO methods are known to exhibit slow convergence and lower accuracy due to high variance in the ZO gradient estimates~\cite{SijiaLiu19}.
While the memory cost is brought down to that of inference, there is an opportunity to further improve memory-efficiency by replacing floating-point arithmetic with low-bit integer arithmetic, which is yet to be fully explored in the literature.

In this paper, we propose a ZO-based on-device learning (ODL) method and its 8-bit version, namely {\MethodName} and {\MethodNameInt}.
{\MethodName} enables full-model training with almost the same memory cost as inference, whereas {\MethodNameInt} trains 8-bit quantized DNNs using only integers.
The contributions of this paper are summarized as follows:
\begin{enumerate}[leftmargin=*]
  \item Unlike pure ZO-based approach, {\MethodName} trains most part of the DNN with ZO but uses BP for the last few layers.
  To the best of our knowledge, this work is the first to consider such a hybrid approach combining ZO with BP.

  \item Building upon {\MethodName}, we propose {\MethodNameInt} for training 8-bit quantized DNNs.
  {\MethodNameInt} is the first ZO-based method that only uses standard 8-bit integer arithmetic.
  To accomplish this, we present a novel method to compute quantized ZO gradients based on the integer cross-entropy loss.
  {\MethodNameInt} enables efficient utilization of limited computing resources on edge devices and makes ZO-based training applicable to low-cost devices without access to floating-point units (FPUs).

  \item We evaluate the performance of {\MethodName} and {\MethodNameInt} on image and point cloud classification datasets.
  {\MethodName} achieves a fast convergence speed and high accuracy comparable to BP, with almost the same memory usage and execution time as ZO (5.2--9.5\% improvements over vanilla ZO with 0.072--1.7\% memory overhead), thereby demonstrating a better compromise between accuracy and training cost.
  {\MethodNameInt} further reduces memory cost and accelerates the training process (1.46--1.60x memory savings and 1.38--1.42x speedup) without degrading the accuracy.
  Experimental results also show that {\MethodName} can be applied to fine-tuning as well as training from initialization.
\end{enumerate}


\section{Related Work} \label{sec:related}
\textbf{Zeroth-order optimization}: There exists a rich body of work on ZO-based methods for BP-free training~\cite{YuriiNesterov17,SaeedGhadimi13,XiangruLian16,SijiaLiu18B,KaiyiJi19,SijiaLiu19,SijiaLiu20,YihuaZhang24}.
ZO-SGD~\cite{SaeedGhadimi13} is a seminal work that replaces a FO gradient in an SGD update rule with the ZO counterpart.
Since ZO-SGD, many ZO methods have been proposed by taking ideas from popular FO methods~\cite{DiederikPKingma15,SashankJReddi16,JeremyBernstein18}.
ZO-SVRG~\cite{SijiaLiu18B,KaiyiJi19} aims to reduce a variance in the ZO gradient using a full gradient computed over the entire dataset, while this incurs a substantial computational overhead.
To address noise in ZO gradients, ZO-signSGD~\cite{SijiaLiu19} only uses a sign information of the ZO gradient, while Huang \textit{et al.}~\cite{FeihuHuang22}, ZO-AdaMM~\cite{XiangyiChen19}, and ZO-AdaMU~\cite{ShuoranJiang24} introduce momentum-based techniques at the cost of increased memory requirements.

\textbf{ZO methods for fine-tuning tasks}: Recently, a memory-efficient version of ZO optimizer, MeZO~\cite{SadhikaMalladi23}, has catalyzed the adoption of ZO methods in fine-tuning tasks (especially for LLMs) and led to follow-up studies~\cite{TanmayGautam24,YifanYang24,ShuoranJiang24,YongLiu24,WentaoGuo24,XinyuTang24,ZhenQin24,ZhenqingLing24,YihuaZhang24}.
While ZO methods require twice the memory of inference to store DNN parameters as well as a random noise added to them, MeZO has (nearly) the same memory cost as inference, because it only saves random seeds used to generate the noise.
MeZO-SVRG~\cite{TanmayGautam24} incorporates a variance-reduction technique by extending ZO-SVRG~\cite{SijiaLiu18B}, while it requires three times the memory of inference.
AdaZeta~\cite{YifanYang24} employs tensorized adapters as a parameter-efficient fine-tuning (PEFT) method, whereas Sparse-MeZO~\cite{YongLiu24} and Guo \textit{et al.}~\cite{WentaoGuo24} only fine-tune a subset of parameters for speedup and improved memory-efficiency.
Tang \textit{et al.}~\cite{XinyuTang24} explores a differentially-private LLM fine-tuning, whereas several works apply MeZO to the federated setting~\cite{ZhenQin24,ZhenqingLing24}.
Since memory-efficiency and algorithmic simplicity of MeZO are attractive for embedded platforms with limited resources, ZO-based on-device learning is emerging as a new research topic~\cite{YequanZhao24,DanPeng24}.

\textbf{ZO methods for on-device learning}: Zhao \textit{et al.}~\cite{YequanZhao24} develops a ZO-based training framework for microcontrollers (MCUs).
It tackles slow convergence of ZO optimization by employing learning-rate scheduling and dimension reduction techniques.
While requiring tens of forward passes per input and taking considerably longer ($\sim$30x) training time than that of BP, their method achieves an accuracy comparable to BP, demonstrating that ZO training is a promising approach under the limited memory budget.
Unlike pure ZO methods, this work explores a different approach where we train DNNs based on ZO but with a little help from BP (FO gradients), which effectively closes the accuracy gap between ZO and BP methods with a negligible memory and execution time overhead.
PocketLLM~\cite{DanPeng24} runs LLM fine-tuning based on MeZO and Adam using off-the-shelf smartphones.
While MeZO is able to perform fine-tuning with a 8x larger batch size than Adam, the implementation is not well optimized for mobile edge devices without GPUs.
Compared to that, we design a ZO method for 8-bit quantized models that achieve a comparable performance to full-precision training and implement it on a low-cost edge device (Raspberry Pi Zero 2).
Notably, the proposed {\MethodNameInt} exclusively uses standard 8-bit integer arithmetic, which allows for fully exploiting computing capability of edge devices and for deploying on-device learning on the platform without FPUs.


\section{Preliminaries} \label{sec:prelim}
\subsection{Zeroth-Order Optimization} \label{sec:prelim-zo}
ZO optimization provides a much simpler way to train DNNs compared to using BP and gradient descent (i.e., FO gradient-based optimization).
To illustrate the idea of ZO-based training, let us consider an $L$-layer network $\vb*{\theta} = \left\{ \vb*{\theta}_1, \ldots, \vb*{\theta}_L \right\}$ and a loss function $\mathcal{L}(\vb*{\theta}; \mathcal{B})$ to be minimized, where $\vb*{\theta}_l$ denotes parameters of the $l$-th layer (concatenated into a single vector) and $\mathcal{B}$ a minibatch taken from a dataset $\mathcal{D}$.
A minibatch $\mathcal{B} = (\mathcal{X}, \mathcal{Y})$ consists of a pair of inputs and their corresponding labels.
BP-based training involves the following four steps: forward pass, loss computation, backward pass, and parameter update (gradient descent).
In the backward pass, a FO parameter gradient $\vb*{g}_l = \nabla_{\vb*{\theta}_l} \mathcal{L}$ is computed from an error $\vb*{e}_l = \nabla_{\vb*{a}_l} \mathcal{L}$ (passed from the next layer) and an activation $\vb*{a}_l$ (computed during the forward pass) using chain-rule.
On the other hand, in ZO methods, a parameter gradient $\hat{\nabla}_{\vb*{\theta}} \mathcal{L}$ is obtained by a ZO gradient estimator, namely Stochastic Perturbation Stochastic Approximation (SPSA), as follows~\cite{JCSpall92,SadhikaMalladi23,YihuaZhang24}:
\begin{equation}
  \hat{\nabla}_{\vb*{\theta}} \mathcal{L}(\vb*{\theta}; \mathcal{B}) =
    \frac{\mathcal{L}(\vb*{\theta} + \varepsilon \vb*{z}; \mathcal{B}) -
    \mathcal{L}(\vb*{\theta} - \varepsilon \vb*{z}; \mathcal{B})}{2 \varepsilon} \vb*{z} = g \vb*{z},
  \label{eq:spsa}
\end{equation}
where $\vb*{z} \sim \mathcal{N}(\vb{0}, \vb{I})$ denotes a Gaussian noise perturbation and $\varepsilon \in \mathbb{R}_+$ is referred to as a perturbation scale.
More simply, a ZO gradient $\hat{\nabla}_{\vb*{\theta}} \mathcal{L}$ is obtained by scaling a random vector $\vb*{z}$ with a scalar quantity $g$.
The projected gradient $g$ is based on the difference between two loss values computed using slightly modified parameters $\vb*{\theta} \pm \varepsilon \vb*{z}$.
Note that $g$ represents a finite approximation of the directional derivative $\vb*{z}^\top \nabla_{\vb*{\theta}} \mathcal{L}$, and therefore $g \vb*{z}$ converges to $\vb*{z} \vb*{z}^\top \nabla_{\vb*{\theta}} \mathcal{L}$ in the limit $\varepsilon \to 0$.
In addition, since $\vb*{z}$ is sampled from a normal distribution, it turns out that $\mathbb{E}[g \vb*{z}] = \mathbb{E}[\vb*{z} \vb*{z}^\top \nabla_{\vb*{\theta}} \mathcal{L}] = \mathbb{E}[\nabla_{\vb*{\theta}} \mathcal{L}]$ holds as $\varepsilon \to 0$ ($\mathbb{E}[\vb*{z} \vb*{z}^\top] = \vb{I}$).
This indicates that Eq. \ref{eq:spsa} (SPSA) gives an unbiased estimation of the FO gradient $\nabla_{\vb*{\theta}} \mathcal{L}$.
The layer parameters $\vb*{\theta}_l$ are updated by gradient descent $\vb*{\theta}_l \gets \vb*{\theta}_l - \eta g \vb*{z}_l$, where $\vb*{z}_l \in \vb*{z}$ denotes a random perturbation for $\vb*{\theta}_l$ and $\eta \in \mathbb{R}_+$ a learning rate.
While BP-based training requires one forward and one backward pass, ZO-based training involves two forward passes per minibatch.

BP methods need to store four types of variables on memory: $\vb*{\theta}, \vb*{a}, \vb*{g}, \vb*{e}$, and therefore consume twice the memory of inference.
On the other hand, ZO methods are inherently memory-efficient as they do not backpropagate errors and only need to keep three variables: $\vb*{\theta}, \vb*{a}, \vb*{z}$.
Besides, in ZO methods, activation tensors $\vb*{a}$ can be immediately released after the forward pass is complete, as they are unnecessary for computing $\hat{\nabla}_{\vb*{\theta}} \mathcal{L}$.
Compared to that, BP methods need to retain all layer activations to compute gradients and errors in the backward pass.
Since ZO methods are BP-free, existing inference engines can be directly used to accelerate the training process as well.

\subsection{MeZO: Memory-Efficient Zeroth-Order Optimizer} \label{sec:prelim-mezo}
While ZO methods consume less memory than BP methods, they still need extra buffers for a random vector $\vb*{z}$, since the same $\vb*{z}$ should be used three times (when perturbing and updating parameters: $\vb*{\theta} \pm \varepsilon \vb*{z}, \ \vb*{\theta} \gets \vb*{\theta} - \eta g \vb*{z}$).
Since $\vb*{z}$ is of the same size as $\vb*{\theta}$ (or $\vb*{g}$), the memory usage of vanilla ZO equals to the size of activations ($\vb*{a}$) plus \textbf{twice} the size of parameters ($\vb*{\theta}, \vb*{z}$), which may hinder training of large-scale networks.
To address this problem, MeZO~\cite{SadhikaMalladi23} proposes a simple yet effective memory reduction trick.
MeZO saves a random seed $s$ instead of a random vector $\vb*{z}$, and resets a pseudorandom generator with $s$ as necessary to reproduce the same $\vb*{z}$.
Since this trick eliminates the need to save $\vb*{z}$ and a random seed is just a 4-byte integer, the memory consumption of MeZO is nearly the same as that of inference (i.e., dominated by $\vb*{\theta}$ and $\vb*{a}$).
MeZO therefore requires half the memory of BP-based training.
Considering these, MeZO is an attractive method for training on edge devices with limited memory.
Refer to the original paper~\cite{SadhikaMalladi23} for a detailed discussion and theoretical analysis on the convergence.


\section{Method} \label{sec:method}
In this section, we propose {\MethodName} as an memory-efficient method for ZO-based ODL.
The training algorithm is summarized in Alg. \ref{alg:halfzo}.

\begin{figure}[htbp]
  \centering
  \includegraphics[keepaspectratio, width=0.6\linewidth]{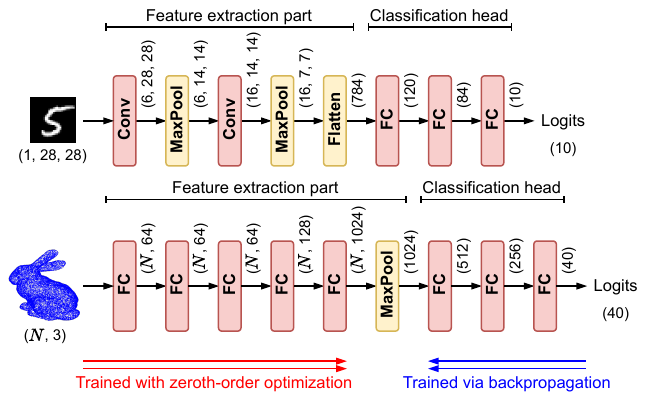}
  \caption{Overview of {\MethodName} (top: LeNet-5, bottom: PointNet).
  {\MethodName} trains the first $C$ layers with ZO optimization and the last $L - C$ layers with BP.
  \textcolor{red}{Red} and \textcolor{Goldenrod}{yellow} rectangles denote layers with and without trainable parameters.}
  \label{fig:half-zo}
\end{figure}

\begin{algorithm}[h]
  \caption{Training with {\MethodName}}
  \label{alg:halfzo}
  \begin{algorithmic}[1]
    \Require Network $\vb*{\theta}$, partition point $C$, steps $T$, batch size $B$,
      perturbation scale $\varepsilon$, learning rate $\eta$
    \For{$t = 0, \ldots, T$}
      \State Sample a minibatch $\mathcal{B}$ (data $\mathcal{X}$ and label $\mathcal{Y}$) from a dataset $\mathcal{D}$
      \State Sample a random seed $s$
      \State \Call{PerturbParameters}{$\vb*{\theta}$, $C$, $s$, $+1$, $\varepsilon$}
        \label{alg:halfzo-perturb0}
      \State Run forward pass and compute a loss: $\ell_+ = \mathcal{L}(\vb*{\theta}; \mathcal{B})$
        \label{alg:halfzo-forward0}
      \State \Call{PerturbParameters}{$\vb*{\theta}$, $C$, $s$, $-2$, $\varepsilon$}
        \label{alg:halfzo-perturb1}
      \State Run forward pass and compute a loss: $\ell_- = \mathcal{L}(\vb*{\theta}; \mathcal{B})$
        \label{alg:halfzo-forward1}
      \State Compute a zeroth-order gradient:
        $g \gets \left( \ell_+ - \ell_- \right) / 2 \varepsilon$
        \label{alg:halfzo-zo-grad}
      \State \Call{PerturbParameters}{$\vb*{\theta}$, $C$, $s$, $+1$, $\varepsilon$}
        \label{alg:halfzo-perturb2}
      \State \Call{ZOUpdateParameters}{%
        $\vb*{\theta}$, $C$, $s$, $\eta$, $g$}
        \label{alg:halfzo-zo-update}
      \State \Call{BPUpdateParameters}{%
        $\vb*{\theta}$, $C$, $(\vb*{a}_C, \ldots, \vb*{a}_L)$, $\mathcal{Y}$}
        \label{alg:halfzo-bp-update}
    \EndFor
    \vspace*{5pt}
    \Function{PerturbParameters}{$\vb*{\theta}$, $C$, $s$, $k$, $\varepsilon$}
      \State Initialize a pseudorandom number generator with a seed $s$
      \For{$l = 1, \ldots, C$}
        \State Generate a random vector:
          $\vb*{z}_l \sim \mathcal{N}(\vb{0}, \vb{I})$
        \State Perturb parameters in-place:
          $\vb*{\theta}_l \gets \vb*{\theta}_l + k \varepsilon \vb*{z}_l$
      \EndFor
    \EndFunction
    \vspace*{5pt}
    \Function{ZOUpdateParameters}{$\vb*{\theta}$, $C$, $s$, $\eta$, $g$}
      \State Initialize a pseudorandom number generator with a seed $s$
      \For{$l = 1, \ldots, C$}
        \State Generate a random vector:
          $\vb*{z}_l \sim \mathcal{N}(\vb{0}, \vb{I})$
        \State Update parameters in-place:
          $\vb*{\theta}_l \gets \vb*{\theta}_l - \eta g \vb*{z}_l$
      \EndFor
    \EndFunction
    \vspace*{5pt}
    \Function{BPUpdateParameters}{$\vb*{\theta}$, $C$, $\vb*{a}$, $\mathcal{Y}$}
      \State Compute a gradient of last layer output $\vb*{e}_L$ using $\vb*{a}_L$ and $\mathcal{Y}$
      \For{$l = L, L - 1, \ldots, C + 1$}
        \State Compute a gradient of parameters $\vb*{g}_l$
          using $\vb*{a}_{l - 1}$ and $\vb*{e}_l$
        \State Compute a gradient of previous layer output $\vb*{e}_{l - 1}$
          using $\vb*{\theta}_l$ and $\vb*{e}_l$
      \EndFor
      \State Update parameters $\vb*{\theta}_{C + 1}, \ldots, \vb*{\theta}_L$ via a FO optimizer (e.g., SGD)
    \EndFunction
  \end{algorithmic}
\end{algorithm}

Let $f(\vb*{\theta})$ denote an $L$-layer DNN with layer parameters $\vb*{\theta} = \left\{ \vb*{\theta}_1, \ldots, \vb*{\theta}_L \right\}$.
As illustrated in Fig. \ref{fig:half-zo}, {\MethodName} splits the given network into two parts, $\left\{ \vb*{\theta}_1, \ldots, \vb*{\theta}_C \right\}$ and $\left\{ \vb*{\theta}_{C + 1}, \ldots, \vb*{\theta}_L \right\}$, and trains them via ZO and BP, respectively.
{\MethodName} can be viewed as an intermediate method between ZO and BP, and {\MethodName} with $C = 0$ and $C = L$ is equivalent to pure BP- and ZO-based training, respectively.
We refer to these cases as \textbf{Full BP} and \textbf{Full ZO}.
While inference engines are optimized for common operations needed during forward passes, such as matrix multiplication and convolution, they could still be used for backward passes as well, considering that backward passes involve similar types of operations (e.g., fully-connected (FC) layers perform matrix multiplications in both forward and backward passes).

As shown in Alg. \ref{alg:halfzo}, {\MethodName} perturbs network parameters using in-place operations (lines \ref{alg:halfzo-perturb0}, \ref{alg:halfzo-perturb1}) and runs forward passes twice to compute losses $\ell_+, \ell_-$ (lines \ref{alg:halfzo-forward0}, \ref{alg:halfzo-forward1}).
These losses are used to obtain a gradient estimate $g$ (line \ref{alg:halfzo-zo-grad}).
{\MethodName} then updates parameters of the first $C$ layers using the estimated gradient $g$ via ZO (line \ref{alg:halfzo-zo-update}) and those of the last $L - C$ layers via BP (line \ref{alg:halfzo-bp-update}).
For BP, {\MethodName} keeps layer activations of the last $L - C + 1$ layers $(\vb*{a}_C, \ldots, \vb*{a}_L)$ obtained during the computation of $\ell_+$ and $\ell_-$.
It is also possible to run another forward pass with the original unperturbed parameters to obtain layer activations, while this approach would require three forward passes per minibatch.
Note that ZO parameter perturbation and update (lines \ref{alg:halfzo-perturb2}--\ref{alg:halfzo-zo-update}) are merged into one step for efficiency, i.e., parameters are updated as $\vb*{\theta}_l \gets \vb*{\theta}_l + (\varepsilon - \eta g) \vb*{z}_l \ (l \le C)$.

\subsection{Memory-Efficiency of {\MethodName}} \label{sec:method-memory-eff}
The memory consumption of {\MethodName} depends on the partition point $C$ and lies in the middle between Full ZO and Full BP.
Full BP requires to keep parameters, layer activations, and their gradients on the memory, and therefore the total memory cost is written as:
\begin{equation}
  M_\text{Full-BP} = \sum_{l \in \mathcal{T}} \left(
    \left| \vb*{\theta}_l \right| + \left| \vb*{g}_l \right| \right)
    + \sum_{l = 1}^L \left( \left| \vb*{a}_l \right| + \left| \vb*{e}_l \right| \right),
  \label{eq:memory-cost-full-bp}
\end{equation}
where $\mathcal{T} \subseteq \left\{ 1, \ldots, L \right\}$ denotes a set of layers with learnable parameters (i.e., FC and convolution layers, \textcolor{red}{red} rectangles in Fig. \ref{fig:half-zo}).
Note that the memory cost can be reduced by considering the lifetime of each variable (e.g., an error $\vb*{e}_l$ can be discarded after $\vb*{g}_l$ and $\vb*{e}_{l - 1}$ are computed).
For ease of analysis, we assume that such optimization is not applied here and thus buffers for all necessary variables remain allocated on memory during the whole training process.
In Full ZO ($C = L$), no extra buffer is needed for gradients and errors as parameters can be updated in-place, and only parameters and layer activations are stored on memory as a result.
The total memory consumption becomes:
\begin{equation}
  M_\text{Full-ZO} = \sum_{l \in \mathcal{T}} \left| \vb*{\theta}_l \right|
    + \sum_{l = 1}^L \left| \vb*{a}_l \right|.
  \label{eq:memory-cost-full-zo}
\end{equation}
While layer activations $\vb*{a}$ can be immediately released after being used by subsequent layers during forward passes, such dynamic memory management is not considered as above to simplify the analysis.
Eqs. \ref{eq:memory-cost-full-bp}--\ref{eq:memory-cost-full-zo} indicate that Full BP takes twice the memory for inference due to gradients $\vb*{g}, \vb*{e}$, while Full ZO only consumes the same amount of memory as inference and thus halves the memory footprint than Full BP.
{\MethodName} stores parameters and activations, as well as gradients of the last layers for BP.
The total memory consumption is given by:
\begin{equation}
  M_\text{{\MethodName}} = \sum_{l \in \mathcal{T}} \left| \vb*{\theta}_l \right|
    + \sum_{l = 1}^L \left| \vb*{a}_l \right|
    + \sum_{\substack{l \in \mathcal{T} \\ l \ge C + 1}} \left| \vb*{g}_l \right|
    + \sum_{l = C + 1}^L \left| \vb*{e}_l \right|.
  \label{eq:memory-cost-halfzo}
\end{equation}
This leads to an inequality $M_\text{Full-ZO} \le M_\text{{\MethodName}} \le M_\text{Full-BP}$, where the equality holds if $C = 0, L$.
The partition point $C$ controls the trade-off between memory usage and accuracy.
Reducing $C$ and training more parts of the network with BP will yield an improved accuracy at the cost of higher memory usage ($M_\text{{\MethodName}}$ approaches to $M_\text{Full-BP}$).
Evaluation results (Sec. \ref{sec:eval}) empirically show that {\MethodName} with $C = L - 1, L - 2$ (i.e., training the last one or two layers with BP and the others with ZO) achieves an accuracy similar to Full BP, while maintaining memory consumption close to that of Full ZO.

The memory consumption depends on the choice of optimizer as well.
If an optimizer does not have any states (e.g., SGD), the memory consumption can be obtained from Eqs. \ref{eq:memory-cost-full-bp}--\ref{eq:memory-cost-halfzo}.
More advanced optimizers consume additional memory to save optimizer states.
For instance, Adam~\cite{DiederikPKingma15} needs to keep track of first- and second-order gradient moments that take up twice the memory of trainable parameters.
In such case, the memory consumption of Full BP is computed as:
\begin{equation}
  M_\text{Full-BP-Adam} = M_\text{Full-BP} + 2\sum_{l \in \mathcal{T}} \left| \vb*{g}_l \right|.
\end{equation}

\subsection{INT8 Training with {\MethodNameInt}} \label{sec:method-int8}
To further improve computational and memory-efficiency, we propose {\MethodNameInt} as a variant of {\MethodName} that only involves 8-bit integer arithmetic.
{\MethodNameInt} integrates the recently-proposed framework for 8-bit integer training, NITI~\cite{MaolinWang22}, with ZO gradient estimation.
In NITI, variables (e.g., parameters and activations) are represented in the form of $\vb*{v}^\text{int8} \cdot 2^s$ and hence stored as a pair of 8-bit signed integers and a scaling exponent $(\vb*{v}^\text{int8}, s)$.
The 8-bit integers are scaled by $2^s$, and hence the exponent is crucial for dynamically adjusting the representable value range during training.

\begin{algorithm}[h]
  \caption{Training with {\MethodNameInt}}
  \label{alg:halfzo-int8}
  \begin{algorithmic}[1]
    \Require Network $\vb*{\theta}^\text{int8}$, partition point $C$, steps $T$, batch size $B$,
      perturbation scale $r_{\max}$, sparsity $p_\text{zero}$, gradient bitwidth $b_\text{ZO}$ (for ZO), $b_\text{BP}$ (for BP)
    \For{$t = 0, \ldots, T$}
      \State Sample a batch $\mathcal{B}$ (data $\mathcal{X}$ and label $\mathcal{Y}$) from a dataset $\mathcal{D}$
      \State Sample a random seed $s$
      \State \Call{PerturbParametersInt8}{%
        $\vb*{\theta}^\text{int8}$, $C$, $s$, $+1$, $r_{\max}$, $p_\text{zero}$}
        \label{alg:halfzo-int8-perturb0}
      \State Run forward pass and compute 8-bit integer logits:
        $\vb*{a}_L^{\text{int8}, +} = f(\vb*{\theta}^\text{int8})$
        \label{alg:halfzo-int8-forward0}
      \State \Call{PerturbParametersInt8}{%
        $\vb*{\theta}^\text{int8}$, $C$, $s$, $-2$, $r_{\max}$, $p_\text{zero}$}
        \label{alg:halfzo-int8-perturb1}
      \State Run forward pass and compute 8-bit integer logits:
        $\vb*{a}_L^{\text{int8}, -} = f(\vb*{\theta}^\text{int8})$
        \label{alg:halfzo-int8-forward1}
      \State Compute a zeroth-order gradient:
        $g \gets \sgn(\ell_+ - \ell_-)
        = \sgn(\mathcal{L}(\vb*{a}_L^{\text{int8}, +}, \mathcal{B}) -
        \mathcal{L}(\vb*{a}_L^{\text{int8}, -}, \mathcal{B}))$
        \label{alg:halfzo-int8-zo-grad}
      \State \Call{PerturbParametersInt8}{%
        $\vb*{\theta}^\text{int8}$, $C$, $s$, $+1$, $r_{\max}$, $p_\text{zero}$}
        \label{alg:halfzo-int8-perturb2}
      \State \Call{ZOUpdateParametersInt8}{%
        $\vb*{\theta}^\text{int8}$, $C$, $s$, $g$, $b_\text{ZO}$}
        \label{alg:halfzo-int8-zo-update}
      \State \Call{BPUpdateParametersInt8}{%
        $\vb*{\theta}^\text{int8}$, $C$,
        $(\vb*{a}_C^\text{int8}, \ldots, \vb*{a}_L^\text{int8})$, $\mathcal{Y}$}
        \label{alg:halfzo-int8-bp-update}
    \EndFor
    \vspace*{5pt}
    \Function{PerturbParametersInt8}{$\vb*{\theta}$, $C$, $s$, $k$, $r_{\max}$, $p_\text{zero}$}
      \label{alg:halfzo-int8-perturb-begin}
      \State Initialize a pseudorandom number generator with a seed $s$
      \For{$l = 1, \ldots, C$}
        \State Generate a random mask and vector:
          $\vb*{m} \sim \text{Bernoulli}(1 - p_\text{zero}), \
          \vb*{u}^\text{int8} \sim \mathcal{U}(-r_{\max}, r_{\max})$
        \State Create a sparse random vector:
          $\vb*{z}_l^\text{int8} \gets \vb*{m} \odot \vb*{u}^\text{int8}$
        \State Perturb parameters in-place:
          $\vb*{\theta}_l^\text{int8} \gets \text{Clamp}(
            \vb*{\theta}_l^\text{int8} + k \vb*{z}_l^\text{int8}, -127, 127)$
          \label{alg:halfzo-int8-perturb-end}
      \EndFor
    \EndFunction
    \vspace*{5pt}
    \Function{ZOUpdateParametersInt8}{$\vb*{\theta}$, $C$, $s$, $g$, $b_\text{ZO}$}
      \label{alg:halfzo-int8-zo-update-begin}
      \State Initialize a pseudorandom number generator with a seed $s$
      \For{$l = 1, \ldots, C$}
        \State Generate a random mask and vector:
          $\vb*{m} \sim \text{Bernoulli}(1 - p_\text{zero}), \
          \vb*{u}^\text{int8} \sim \mathcal{U}(-r_{\max}, r_{\max})$
        \State Create a sparse random vector:
          $\vb*{z}_l^\text{int8} \gets \vb*{m} \odot \vb*{u}^\text{int8}$
        \State Compute an update:
          $\vb*{g}^\text{int8} \gets \text{PseudoStochasticRound}(
            g \vb*{z}_l^\text{int8}, b_\text{ZO})$
        \State Update parameters in-place:
          $\vb*{\theta}_l^\text{int8} \gets \text{Clamp}(
            \vb*{\theta}_l^\text{int8} - \vb*{g}^\text{int8}, -127, 127)$
          \label{alg:halfzo-int8-zo-update-end}
      \EndFor
    \EndFunction
  \end{algorithmic}
\end{algorithm}

Unlike {\MethodName}, {\MethodNameInt} performs each step in Alg. \ref{alg:halfzo} entirely with integer arithmetic.
The resulting algorithm is outlined in Alg. \ref{alg:halfzo-int8}.
Let $f(\vb*{\theta}^\text{int8})$ denote an 8-bit quantized network with $\vb*{\theta}^\text{int8} = \left\{ (\vb*{\theta}_l^\text{int8}, s_{\theta, l}) \right\}_{l = 1}^L$ being layer parameters (8-bit integers) and their associated exponents (scalars).
Similar to {\MethodName}, {\MethodNameInt} first runs two forward passes (lines \ref{alg:halfzo-int8-forward0}, \ref{alg:halfzo-int8-forward1}) using perturbed parameters (lines \ref{alg:halfzo-int8-perturb0}, \ref{alg:halfzo-int8-perturb1}) and keeps outputs of the last layer $\vb*{a}_L^{\text{int8}, +}, \vb*{a}_L^{\text{int8}, -}$.
It computes a zeroth-order gradient $g$ using these logits (line \ref{alg:halfzo-int8-zo-grad}) and restores the unperturbed parameters (line \ref{alg:halfzo-int8-perturb2}).
The parameters of the first $C$ layers are updated with ZO (line \ref{alg:halfzo-int8-zo-update}) and those of the last $L - C$ layers with BP (line \ref{alg:halfzo-int8-bp-update}), respectively.

\textbf{Parameter perturbation} (lines \ref{alg:halfzo-int8-perturb-begin}--\ref{alg:halfzo-int8-perturb-end}): In Full ZO and {\MethodName}, parameters are perturbed with a standard Gaussian random vector $\vb*{z} \sim \mathcal{N}(\vb{0}, \vb{I})$.
{\MethodNameInt} instead samples a random INT8 vector $\vb*{u}^\text{int8}$ from a uniform distribution $\mathcal{U}(-r_{\max}, r_{\max})$, since NITI adopts a uniform initialization of parameters to achieve better accuracy with a limited value range~\cite{MaolinWang22}.
In addition, the vector $\vb*{u}^\text{int8}$ is made sparse by randomly setting $p_\text{zero}$\% of its entries to zeros using a binary mask $\vb*{m} \sim \text{Bernoulli}(1 - p_\text{zero})$.
This can be viewed as a random dropout in standard BP training.
The resulting sparse perturbation $\vb*{z}^\text{int8} = \vb*{m} \odot \vb*{u}^\text{int8}$ is applied to parameters $\vb*{\theta}_l^\text{int8}$ and they are clipped to the INT8 range.

\textbf{Parameter update via ZO} (lines \ref{alg:halfzo-int8-zo-update-begin}--\ref{alg:halfzo-int8-zo-update-end}): Given a learning rate $\eta$ and zeroth-order gradient $g$, {\MethodName} computes an update to the parameter as $-\eta g \vb*{z}$.
Following NITI, {\MethodNameInt} first computes a product $g \vb*{z}^\text{int8}$ and rounds it to $b_\text{ZO}$-bit using a pseudo-stochastic rounding scheme (refer to \cite{MaolinWang22} for details of rounding) to obtain an update $\vb*{g}^\text{int8}$.
The bitwidth $b_\text{ZO}$ determines a magnitude of an update and hence works as a learning rate.
With a scaling exponent $s_\theta$ for parameters $\vb*{\theta}^\text{int8}$, the update is actually written as $\vb*{g}^\text{int8} \cdot 2^{s_\theta}$ ($s_\theta$ is fixed throughout the training process as in NITI).
Note that parameters can be updated in-place and thus no extra memory is required for intermediate variables ($\vb*{m}, \vb*{u}^\text{int8}, \vb*{z}_l^\text{int8}, \vb*{g}^\text{int8}$).

\textbf{Forward and backward passes} (lines \ref{alg:halfzo-int8-forward0}, \ref{alg:halfzo-int8-forward1}, \ref{alg:halfzo-int8-bp-update}): Forward and backward passes follow the NITI framework.
In the forward pass, an input is first converted to the form of $\vb*{a}_0 \cdot 2^{s_{a, 0}}$ before fed to the network.
A FC or convolution layer performs matrix product or convolution between parameters $\vb*{\theta}_l^\text{int8}$ and an input $\vb*{a}_{l - 1}^\text{int8}$ in integer arithmetic and stores the result $\vb*{a}_l^\text{int32}$ in 32-bit integer format.
Besides, a scaling exponent of an output is obtained by adding exponents of the parameters and input, i.e., $s_{a, l} = s_{\theta, l} + s_{a, l - 1}$.
The output is rounded to 8-bit $\vb*{a}_l^\text{int8}$ and the exponent is updated accordingly.
In this rounding process, the minimum bitwidth required to represent values in $\vb*{a}_l^\text{int32}$ is determined as $b = \lfloor \log_2(\max(\abs{\vb*{a}_l^\text{int32}})) \rfloor + 1$, where $\max(\abs{\vb*{a}_l^\text{int32}})$ denotes the maximum absolute value in $\vb*{a}_l^\text{int32}$.
In case of $b > 7$, values in $\vb*{a}_l^\text{int32}$ is right-shifted by $b - 7$-bit and $b - 7$ is added to the exponent $s_{\theta, l}$.
Refer to \cite{MaolinWang22} for a rounding procedure and backward pass.
Similar to ZO, an update to the parameter is rounded to $b_\text{BP}$-bit.

\subsection{Zeroth-order Gradient Estimation with Integer Arithmetic} \label{sec:method-int8-gradient}
In ZO training, the zeroth-order gradient $g$ is derived from the difference between two cross-entropy losses $\ell_+, \ell_-$.
NITI only proposes an integer arithmetic-only method to approximate a gradient of the integer cross-entropy loss $\mathcal{L}(\vb*{a}_L^\text{int8})$ with respect to its logits $\vb*{a}_L^\text{int8}$, and circumvents an evaluation of the loss value itself.
$\mathcal{L}$ involves a log-sum-exp function and seems challenging to compute it using integer arithmetic only.
In addition, since an update $\vb*{g}^\text{int8}$ is obtained by rounding a product $g \vb*{z}$ to $b_\text{ZO}$-bit, the magnitude of zeroth-order gradient $g$ does not affect $\vb*{g}^\text{int8}$ (e.g., $\vb*{z}$, $2\vb*{z}$, and $4\vb*{z}$ are likely to produce the same update vector after rounding).
Considering these, we propose to only evaluate a sign of difference between two losses $\sgn(\ell_+ - \ell_-)$ and use it as a ternary gradient $g \in \left\{ -1, 0, +1 \right\}$.
Besides, losses $\ell_+, \ell_-$ can be computed using floating-point as a simple workaround.

Let $(\vb*{\alpha}^\text{int8}, s_\alpha)$ and $(\vb*{\beta}^\text{int8}, s_\beta)$ denote 8-bit integer logits (i.e., outputs of the network $f(\vb*{\theta}^\text{int8})$) along with their scaling factors.
The cross-entropy loss is written as:
\begin{equation}
  \mathcal{L}(\vb*{\alpha}^\text{int8}; \mathcal{Y}) =
    -\ln \frac{\exp(\alpha_i^\text{int8} \cdot 2^{s_\alpha})}{
      \sum_j \exp(\alpha_j^\text{int8} \cdot 2^{s_\alpha})}, \quad
  \mathcal{L}(\vb*{\beta}^\text{int8}; \mathcal{Y}) =
    -\ln \frac{\exp(\beta_i^\text{int8} \cdot 2^{s_\beta})}{
      \sum_j \exp(\beta_j^\text{int8} \cdot 2^{s_\beta})},
  \label{eq:cross-entropy-loss0}
\end{equation}
where $i$ is a label of the input.
We modify Eq. \ref{eq:cross-entropy-loss0} as:
\begin{equation}
  \mathcal{L}(\vb*{\alpha}^\text{int8}; \mathcal{Y}) =
    \ln \sum_j \exp(\left( \alpha_j^\text{int8} - \alpha_i^\text{int8} \right) \cdot 2^{s_\alpha}), \quad
  \mathcal{L}(\vb*{\beta}^\text{int8}; \mathcal{Y}) =
    \ln \sum_j \exp(\left( \beta_j^\text{int8} - \beta_i^\text{int8} \right) \cdot 2^{s_\beta}).
  \label{eq:cross-entropy-loss1}
\end{equation}
For simplicity, we define a new scaling factor as $s = \min(s_\alpha, s_\beta)$ and rewrite terms in Eq. \ref{eq:cross-entropy-loss1} using rescaled logits $\bar{\alpha}_j^\text{int8} = \alpha_j^\text{int8} \cdot 2^{s_\alpha - s}$, $\bar{\beta}_j^\text{int8} = \beta_j^\text{int8} \cdot 2^{s_\beta - s}$:
\begin{equation}
  \exp(\left( \alpha_j^\text{int8} -
    \alpha_i^\text{int8} \right) \cdot 2^{s_\alpha}) =
    \exp(\left( \bar{\alpha}_j^\text{int8} -
    \bar{\alpha}_i^\text{int8} \right) \cdot 2^s), \
  \exp(\left( \beta_j^\text{int8} -
    \beta_i^\text{int8} \right) \cdot 2^{s_\beta}) =
    \exp(\left( \bar{\beta}_j^\text{int8} -
    \bar{\beta}_i^\text{int8} \right) \cdot 2^s).
\end{equation}
To evaluate the above with integer operations, we replace $\exp(x)$ with a power of two $2^{(\log_2 e) \cdot x}$ and approximate $\log_2 e$ by integer multiplication and bit-shift $\log_2 e \approx 47274 \cdot 2^{-15}$, as discussed in \cite{MaolinWang22}.
This yields the following:
\begin{equation}
  \exp(\left( \bar{\alpha}_j^\text{int8} -
    \bar{\alpha}_i^\text{int8} \right) \cdot 2^s) =
    2^{47274 \left( \bar{\alpha}_j^\text{int8} -
    \bar{\alpha}_i^\text{int8} \right) \cdot 2^{s - 15}} = 2^{\hat{\alpha}_j}, \
  \exp(\left( \bar{\beta}_j^\text{int8} -
    \bar{\beta}_i^\text{int8} \right) \cdot 2^s) =
    2^{47274 \left( \bar{\beta}_j^\text{int8} -
    \bar{\beta}_i^\text{int8} \right) \cdot 2^{s - 15}} = 2^{\hat{\beta}_j}.
\end{equation}
The exponent $\hat{\alpha}_j, \hat{\beta}_j$ may still be large and cause an integer overflow.
For numerical stability, we offset each exponent by $p = p_{\max} - 10$ ($p_{\max} = \max(\max_j(\hat{\alpha}_j), \max_j(\hat{\beta}_j))$) and define a new exponent as $\tilde{\alpha}_j = \max(\hat{\alpha}_j - p, 0), \tilde{\beta}_j = \max(\hat{\beta}_j - p, 0)$, such that $2^{\tilde{\alpha}_j}, 2^{\tilde{\beta}_j} \le 2^{10}$ holds for any $j$~\cite{MaolinWang22}.
The term $2^x$ with an exponent $x < p$ can be safely ignored as it is at least $2^{10}$ times smaller than the maximum term $2^{p_{\max}} = 2^{p + 10}$ and has little effect on the final result.
With these notations, the loss difference can be written as:
\begin{equation}
  \mathcal{L}(\vb*{\alpha}^\text{int8}; \mathcal{Y}) -
    \mathcal{L}(\vb*{\beta}^\text{int8}; \mathcal{Y}) =
    \ln \frac{\sum_j \exp(\left( \alpha_j^\text{int8} -
      \alpha_i^\text{int8} \right) \cdot 2^{s_\alpha})}{
      \sum_j \exp(\left( \beta_j^\text{int8} -
      \beta_i^\text{int8} \right) \cdot 2^{s_\beta})} =
  \ln \frac{\sum_j 2^{\tilde{\alpha}_j}}{\sum_j 2^{\tilde{\beta}_j}}.
  \label{eq:cross-entropy-loss-int8}
\end{equation}
Eq. \ref{eq:cross-entropy-loss-int8} involves evaluating a ratio of sums of terms (approximated as powers of two) and taking its natural logarithm.
Since it seems challenging to obtain an accurate result in a form of $x \cdot 2^s$ using integer arithmetic only, we consider a sign of loss difference and use it as a gradient $g$ (Alg. \ref{alg:halfzo-int8}, line \ref{alg:halfzo-int8-zo-grad}).
The sign can be easily obtained by comparing two terms $\sum_j 2^{\tilde{\alpha}_j}, \sum_j 2^{\tilde{\beta}_j}$ without logarithm operations.
Similarly, ZO-signSGD~\cite{SijiaLiu19} uses a sign information of the zeroth-order gradient estimate to mitigate the effect of gradient noise and improve robustness of ZO training.

If a batch size $B$ is greater than one, the loss difference has the following form:
\begin{equation}
  \mathcal{L}(\vb*{\alpha}^\text{int8}; \mathcal{Y}) -
    \mathcal{L}(\vb*{\beta}^\text{int8}; \mathcal{Y}) =
    \sum_b \ln \frac{\sum_j \exp(\left( \alpha_{b, j}^\text{int8} -
    \alpha_{b, t(b)}^\text{int8} \right) \cdot 2^{s_{\alpha, b}})}{
    \sum_j \exp(\left( \beta_{b, j}^\text{int8} -
    \beta_{b, t(b)}^\text{int8} \right) \cdot 2^{s_{\beta, b}})} =
  \sum_b \ln \frac{\sum_j 2^{\tilde{\alpha}_{b, j}}}{\sum_j 2^{\tilde{\beta}_{b, j}}},
  \label{eq:cross-entropy-loss-int8-batched0}
\end{equation}
where $b$ denotes a sample index in a minibatch and $t(b)$ a label for the $b$-th sample.
In this case, we use Eq. \ref{eq:cross-entropy-loss-int8-batched} to evaluate a sign:
\begin{equation}
  \mathcal{L}(\vb*{\alpha}^\text{int8}; \mathcal{Y}) -
    \mathcal{L}(\vb*{\beta}^\text{int8}; \mathcal{Y}) =
  \ln 2 \cdot \left( \sum_b \lfloor \log_2 \sum_j 2^{\tilde{\alpha}_{b, j}} \rfloor -
    \sum_b \lfloor \log_2 \sum_j 2^{\tilde{\beta}_{b, j}} \rfloor \right).
    \label{eq:cross-entropy-loss-int8-batched}
\end{equation}
Note that $\lfloor \log_2 n \rfloor$ is easily obtained by counting the number of leading zero bits in $n$.
While the use of floor operation $\lfloor \cdot \rfloor$ may lead to incorrect results, we empirically find that correct signs can be obtained at a high probability ($\sim$95\%).

\subsection{Memory-Efficiency of {\MethodNameInt}} \label{sec:method-int8-memory-eff}
{\MethodNameInt} with $C = 0$ (Full BP) is equivalent to the baseline NITI ({\MethodNameInt} with $C = L$ (Full ZO) is a ZO-based 8-bit training algorithm).
While 8-bit integers are used, a FC or convolution layer requires extra buffers to store intermediate multiply-accumulation results (32-bit integers) when computing an activation $\vb*{a}_l$, gradient $\vb*{g}_l$, and error $\vb*{e}_{l - 1}$.
In NITI, the maximum absolute value of an input tensor $\vb*{x}^\text{int32}$ should be obtained first before rounding its elements to 8-bit and adjusting a scaling factor, necessitating a buffer to store the whole input tensor.
As a result, the total memory consumption is formulated as:
\begin{align}
  M_\text{Full-BP-Int8} &= \sum_{l \in \mathcal{T}} \left(
    \left| \vb*{\theta}_l^\text{int8} \right| +
    \left| \vb*{g}_l^\text{int8} \right| \right) +
    \sum_{l = 1}^L \left(
    \left| \vb*{a}_l^\text{int8} \right| +
    \left| \vb*{e}_l^\text{int8} \right| \right) +
    \sum_{l \in \mathcal{T}} \left(
    \left| \vb*{a}_l^\text{int32} \right| +
    \left| \vb*{g}_l^\text{int32} \right| \right) +
    \sum_{\substack{l \in \mathcal{T} \\ l > 1}}
    \left| \vb*{e}_{l - 1}^\text{int32} \right| \nonumber \\
  &= \frac{1}{4} M_\text{Full-BP} +
  \sum_{l \in \mathcal{T}} \left(
    \left| \vb*{a}_l^\text{int32} \right| +
    \left| \vb*{g}_l^\text{int32} \right| \right) +
    \sum_{\substack{l \in \mathcal{T} \\ l > 1}}
    \left| \vb*{e}_{l - 1}^\text{int32} \right|.
  \label{eq:memory-cost-full-bp-int8}
\end{align}
In case of $C = L$ (i.e., all layers are trained via ZO), gradients and errors are not needed to be kept on memory as parameters are updated in-place, leading to the reduced memory usage:
\begin{equation}
  M_\text{Full-ZO-Int8} = \sum_{l \in \mathcal{T}}
    \left| \vb*{\theta}_l^\text{int8} \right| +
    \sum_{l = 1}^L \left| \vb*{a}_l^\text{int8} \right| +
    \sum_{l \in \mathcal{T}} \left| \vb*{a}_l^\text{int32} \right|
  = \frac{1}{4} M_\text{Full-ZO} +
    \sum_{l \in \mathcal{T}} \left| \vb*{a}_l^\text{int32} \right|.
  \label{eq:memory-cost-full-zo-int8}
\end{equation}
This indicates the memory usage is reduced by more than half and equals to the inference.
{\MethodNameInt} stores parameters and activations for all layers, as well as gradients and errors for the last $L - C + 1$ layers, which gives the total memory cost:
\begin{align}
  M_\text{{\MethodNameInt}} &= \sum_{l \in \mathcal{T}}
    \left| \vb*{\theta}_l^\text{int8} \right| +
    \sum_{l = 1}^L \left| \vb*{a}_l^\text{int8} \right| +
    \sum_{\substack{l \in \mathcal{T} \\ l \ge C + 1}}
    \left| \vb*{g}_l^\text{int8} \right| +
    \sum_{l = C + 1}^L \left| \vb*{e}_l^\text{int8} \right| \nonumber \\
  &\phantom{=} +
    \sum_{l \in \mathcal{T}} \left| \vb*{a}_l^\text{int32} \right| +
    \sum_{\substack{l \in \mathcal{T} \\ l \ge C + 1}}
    \left| \vb*{g}_l^\text{int32} \right| +
    \sum_{\substack{l \in \mathcal{T} \\ l > C + 1}}
    \left| \vb*{e}_{l - 1}^\text{int32} \right| \nonumber \\
  &= \frac{1}{4} M_\text{{\MethodName}} +
    \sum_{l \in \mathcal{T}} \left| \vb*{a}_l^\text{int32} \right| +
    \sum_{\substack{l \in \mathcal{T} \\ l \ge C + 1}}
    \left| \vb*{g}_l^\text{int32} \right| +
    \sum_{\substack{l \in \mathcal{T} \\ l > C + 1}}
    \left| \vb*{e}_{l - 1}^\text{int32} \right|.
  \label{eq:memory-cost-halfzo-int8}
\end{align}
Similar to Sec. \ref{sec:method-memory-eff}, this gives an inequality $M_\text{Full-ZO-Int8} \le M_\text{{\MethodNameInt}} \le M_\text{Full-BP-Int8}$, and therefore the memory footprint of {\MethodNameInt} is in the middle between that of Full BP and Full ZO.
In addition, the memory usage of the INT8 version is not equal to a quarter of the FP32 case (Eqs. \ref{eq:memory-cost-full-bp}--\ref{eq:memory-cost-halfzo}) due to additional buffers for intermediate results.
Recall that the memory usage can be further reduced by taking into account the lifetime of variables and reusing the same buffer for variables with a non-overlapping lifetime (e.g., $\vb*{e}_l^\text{int8}$ and $\vb*{e}_{l - 2}^\text{int8}$ in Eq. \ref{eq:memory-cost-full-bp-int8} and $\vb*{a}_l^\text{int32}$ and $\vb*{a}_{l + 2}^\text{int32}$ in Eq. \ref{eq:memory-cost-full-zo-int8}).
Such memory optimization techniques are well-studied and orthogonal to our work.
We plan to improve Eqs. \ref{eq:memory-cost-full-bp}--\ref{eq:memory-cost-halfzo} and Eqs. \ref{eq:memory-cost-full-bp-int8}--\ref{eq:memory-cost-halfzo-int8} in a future work.


\section{Evaluation} \label{sec:eval}
In this section, we evaluate {\MethodName} and {\MethodNameInt} in terms of three aspects: accuracy, memory-efficiency, and execution time.
We compare the performance of proposed methods with Full ZO and Full BP in both FP32 and INT8 cases.

\subsection{Experimental Setup} \label{sec:eval-setup}
{\MethodName}, {\MethodNameInt}, as well as Full BP and Full ZO are implemented with PyTorch.
The INT8 version is based on the publicly available code of NITI~\cite{MaolinWang22}.
For accuracy evaluation, we use a workstation with an eight-core 3.8GHz Intel Core i7-10700K CPU, a 32GB of DRAM, and an Nvidia GeForce RTX 3090 GPU running Ubuntu 20.04.
We conduct experiments on PyTorch 2.5.0 and CUDA 12.1.

In addition, {\MethodName} and {\MethodNameInt} are implemented in C++ and cross-compiled with \texttt{-O3} optimization flag using ARM GCC toolchain 12.2.
We employ vectorization with ARM NEON intrinsics and multithreading with OpenMP library to fully utilize CPU cores.
The training code is executed on a Raspberry Pi Zero 2 board with a four-core 1GHz ARM Cortex-A72 CPU and a 512MB of DRAM running Raspberry Pi OS (based on Debian 12).
Note that the CPU frequency is fixed at 1GHz to accurately measure the execution time.

As for datasets, we use MNIST~\cite{YannLecun98}, Fashion-MNIST~\cite{HanXiao17}, and ModelNet40~\cite{ZhirongWu15,CharlesRQi17} in the evaluation.
The former two are widely-known image classification datasets, each consisting of \num{50000} 28$\times$28 grayscale images for training and \num{10000} images for testing.
On the other hand, ModelNet40~\cite{ZhirongWu15,CharlesRQi17} is a 3D point cloud dataset containing \num{12311} samples from 40 object categories (\num{9843} for training and \num{2468} for testing).
Each sample consists of \num{2048} points that are randomly sampled from a synthetic 3D model and normalized to have a zero centroid and a unit radius.
When evaluating a fine-tuning performance, we additionally use Rotated MNIST and Rotated Fashion-MNIST~\cite{VihariPiratla20,HonokaAnada24}.
We randomly choose \num{1024} images from train and test splits and rotate them by either $30^\circ$ or $45^\circ$.

\subsubsection{Details of DNN Models and Training} \label{sec:eval-setup-training}
We train LeNet-5~\cite{YannLecun98} and PointNet~\cite{CharlesRQi17} on MNIST and ModelNet40 datasets, respectively.
LeNet-5 (Fig. \ref{fig:half-zo}, top) consists of a stack of two convolution layers followed by three FC layers.
In {\MethodName} and {\MethodNameInt}, we consider setting the partition point to $C = L - 1, L - 2$.
That is, we consider training (i) two convolution layers and the first FC layer, or (ii) two convolution layers and the first two FC layers via ZO, and the remaining one or two FC layers via BP.
As a result, 89.8\% and 99.2\% of parameters (\num{96772} and \num{106936} out of \num{107786}) are trained via ZO.
We refer to these configurations as \textbf{ZO-Feat-Cls1} and \textbf{ZO-Feat-Cls2}.
PointNet (Fig. \ref{fig:half-zo}, bottom) is a DNN tailored for point cloud tasks and directly takes 3D point coordinates as input.
The feature extraction part consists of five FC layers and one max-pooling layer, while the classification head is a stack of three FC layers.
Similar to LeNet-5, we train convolution layers as well as the first (i) one or (ii) two FC layers via ZO, meaning that ZO handles 82.7\% and 98.7\% of the parameters (\num{675136} and \num{806464} out of \num{816744}), respectively.
Note that 8-bit models do not have bias parameters as in NITI~\cite{MaolinWang22}.

Throughout the evaluation, we use vanilla SGD optimizer without momentum or weight decay for updating parameters.
The number of epochs is set to $E = 100, 200$ when training LeNet-5 and PointNet from scratch, respectively.
We describe the training details below.

\textbf{FP32 case}:
For each model and dataset, the initial learning rate $\eta$ is tuned in the range $[\text{1e-4}, \text{5e-2}]$.
It is decayed by a factor of 0.8 every 10 epochs.
The batch size is set to $B = 32$.
In ZO training, we clip a ZO gradient $g$ within the range $[-g_\text{clip}, g_\text{clip}]$ to stabilize training.

\textbf{INT8 case}:
Following NITI~\cite{MaolinWang22}, the gradient bitwidth $b_\text{BP}$ (for BP) is first set to 5 and then decreased to 4 and 3 at 20 and 50 epochs, respectively.
The batch size is set to $B = 256$.
In ZO training, the perturbation scale $\varepsilon$ is tuned in the range $[1, 3, 7, 15, 31, 63]$ for each experiment, while the gradient bitwidth $b_\text{ZO}$ is fixed to 1.
The sparsity $p_\text{zero}$ is initialized with 0.33 and increased to 0.5 and 0.9 at 20 and 50 epochs, respectively.

\subsection{Accuracy on Classification Datasets} \label{sec:eval-acc}
Table \ref{tbl:accuracy} (columns 1--6) shows the classification accuracy of LeNet-5 on MNIST and Fashion-MNIST.
INT8 and INT8* denote that ZO gradients $g$ are obtained using floating-point and integer arithmetic, respectively (Sec. \ref{sec:method-int8-gradient}).
As a result, INT8* performs an integer arithmetic-only training.
We evaluate the accuracy of FP32 and INT8 methods using the PyTorch implementation and that of INT8* counterpart using C++ version.

In the FP32 case (columns 1, 4), Full ZO gives 9.3--14.3\% lower accuracy than Full BP, which is attributed to a large noise in ZO gradients $g$.
On the other hand, by training the last one or two FC layers (i.e., 0.8\% and 10.2\% of LeNet-5 parameters) via BP, ZO-Feat-Cls2 and ZO-Feat-Cls1 improve the accuracy by 5.1--5.2\% and 7.7--9.5\% with a negligible execution time and memory overhead (Figs. \ref{fig:memory-usage} and \ref{fig:execution-time}).
As a result, ZO-Feat-Cls2 and ZO-Feat-Cls1 shrink the accuracy gap to 4.3--9.1\% and 1.6--4.8\%, respectively.
In the INT8 case (columns 2--3, 5--6), Full ZO shows 9.0--16.4\% lower accuracy than Full BP.
ZO-Feat-Cls2 and ZO-Feat-Cls1 produce 7.6--10.7\% and 4.6--6.4\% higher accuracy than Full ZO, reducing the accuracy gap to 4.4--10.1\% and 1.4--5.7\%, respectively.
These results indicate that {\MethodName} and {\MethodNameInt} address the low accuracy of Full ZO with a little help from BP.

As seen in columns 1--2 and 4--5, {\MethodNameInt} provides nearly the same or comparable accuracy to that of {\MethodName} while improving memory-efficiency (Fig. \ref{fig:memory-usage-int8}).
The accuracy difference is at most 0.5\% and 2.0\% in MNIST and Fashion-MNIST, respectively.
The proposed integer-only gradient estimation (INT8*, Sec. \ref{sec:method-int8-gradient}) leads to 0.4--1.5\% and 2.4--3.1\% accuracy drop in MNIST and Fashion-MNIST, respectively, due to incorrect estimation results (around 5\% of the time).
Despite that, it still preserves a comparable level of accuracy as the floating-point counterpart as seen in columns 2--3 and 5--6.

The accuracy of PointNet on ModelNet40 is presented in Table \ref{tbl:accuracy} (column 7).
Full ZO fails to achieve satisfactory accuracy when training PointNet (which has 7.6x more parameters than LeNet-5) from random initialization.
Compared to that, {\MethodName} largely outperforms Full ZO and also shows a comparable or even better accuracy to Full BP by training only the last few FC layers via BP.

{\MethodNameInt} achieves better prediction accuracy by gradually increasing a sparsity of parameter updates $p_\text{zero}$ (Alg. \ref{alg:halfzo-int8}, line \ref{alg:halfzo-int8-zo-update}) during training.
For example, when $p_\text{zero}$ is fixed to the initial value (0.33) throughout the entire training process, Full ZO yields 9.5\% and 6.3\% lower accuracy on MNIST and Fashion-MNIST (80.26\%/89.78\% vs. 67.72\%/73.98\%), respectively.

\begin{table}[h]
  \centering
  \caption{Accuracy of LeNet-5 (MNIST and Fashion-MNIST) and PointNet (ModelNet40)}
  \label{tbl:accuracy}
  \begin{tabular}{l|rrr|rrr|r} \hline
    & \multicolumn{3}{c|}{MNIST} & \multicolumn{3}{c|}{Fashion-MNIST} &
      \multicolumn{1}{c}{ModelNet40} \\
    & FP32 & INT8 & INT8* & FP32 & INT8 & INT8* & FP32 \\ \hline
    Full ZO & 89.80 & 89.78 & 88.92 & 77.09 & 73.98 & 71.02 & 32.05 \\
    ZO-Feat-Cls2 & 94.85 & 94.34 & 93.92 & 82.28 & 80.33 & 77.93 & 70.38 \\
    ZO-Feat-Cls1 & 97.53 & 97.34 & 95.83 & 86.60 & 84.66 & 81.60 & 73.50 \\
    Full BP & 99.10 & 98.77 & -- & 91.37 & 90.40 & -- & 71.60 \\ \hline
  \end{tabular}
\end{table}

Table \ref{tbl:accuracy-fine-tuning} shows fine-tuning results on MNIST datasets.
In the FP32 case, LeNet-5 is first pre-trained on MNIST and Fashion-MNIST for a single epoch using BP.
We use an Adam optimizer with an initial learning rate of $\eta = \text{1e-3}$ and momentum parameters of $\beta_1 = 0.9, \beta_2 = 0.999$.
LeNet-5 is then fine-tuned on rotated datasets for additional 50 epochs.
In the INT8 case, LeNet-5 is pre-trained on the original datasets for 100 epochs using BP (i.e., NITI) with a gradually decreasing gradient bit-width $b_\text{BP}$ (Sec. \ref{sec:eval-setup-training}).
In the fine-tuning step, LeNet-5 is trained for 50 more epochs on rotated datasets.

Without fine-tuning, the model suffers from a significant accuracy loss in case of a larger rotation angle ($\theta = 45^\circ$), which is due to the discrepancy between training and test data distributions.
Note that we do not employ data augmentation (e.g., random rotation) during pre-training, i.e., baseline models are not trained with rotated images.
For FP32 models with Rotated MNIST (columns 1--2), fine-tuning with ZO-Feat-Cls2/ZO-Feat-Cls1 leads to 15.6/18.8\% ($\theta = 30^\circ$) and 39.7/45.0\% ($\theta = 45^\circ$) accuracy improvements.
Similarly, in case of Rotated Fashion-MNIST (columns 3--4), fine-tuning with ZO-Feat-Cls2/ZO-Feat-Cls1 improves the prediction accuracy by 37.6/36.3\% and 51.2/53.5\%, respectively.
By training a small fraction (10.2\%) of the network via BP, ZO-Feat-Cls2 and ZO-Feat-Cls1 successfully bridge the accuracy gap between Full ZO and Full BP.
As shown in columns 5--8, while Full ZO is affected by 8-bit quantization especially with Rotated Fashion-MNIST, {\MethodNameInt} achieves an accuracy close to that of the full-precision counterparts.
These results indicate that both {\MethodName} and {\MethodNameInt} can be used for fine-tuning as well as training from scratch and effectively handle the data distribution shift.

\begin{table}[h]
  \centering
  \caption{Accuracy of LeNet-5 w/ or w/o fine-tuning on Rotated MNIST and Rotated Fashion-MNIST}
  \label{tbl:accuracy-fine-tuning}
  \begin{tabular}{l|rr|rr|rr|rr} \hline
    & \multicolumn{4}{c|}{FP32} & \multicolumn{4}{c}{INT8} \\ \cline{2-9}
    & \multicolumn{2}{c|}{Rotated MNIST} & \multicolumn{2}{c|}{Rotated F-MNIST} &
      \multicolumn{2}{c|}{Rotated MNIST} & \multicolumn{2}{c}{Rotated F-MNIST} \\
    & $30^\circ$ & $45^\circ$ & $30^\circ$ & $45^\circ$ &
      $30^\circ$ & $45^\circ$ & $30^\circ$ & $45^\circ$ \\ \hline
    w/o Fine-tuning & 74.41 & 46.58 & 39.65 & 23.34 &
      84.08 & 60.25 & 28.13 & 13.77 \\
    Full ZO & 85.94 & 74.71 & 61.33 & 59.08 &
      85.94 & 64.36 & 37.21 & 22.95 \\
    ZO-Feat-Cls2 & 90.04 & 86.23 & 77.25 & 74.51 &
      93.07 & 87.99 & 69.43 & 68.95 \\
    ZO-Feat-Cls1 & 93.16 & 91.60 & 75.98 & 76.86 &
      93.46 & 91.80 & 76.07 & 75.68 \\
    Full BP & 94.82 & 93.85 & 80.37 & 79.49 &
      96.68 & 95.21 & 80.08 & 80.86 \\ \hline
  \end{tabular}
\end{table}

\begin{figure}[h]
  \centering
  \includegraphics[keepaspectratio, width=0.8\linewidth]{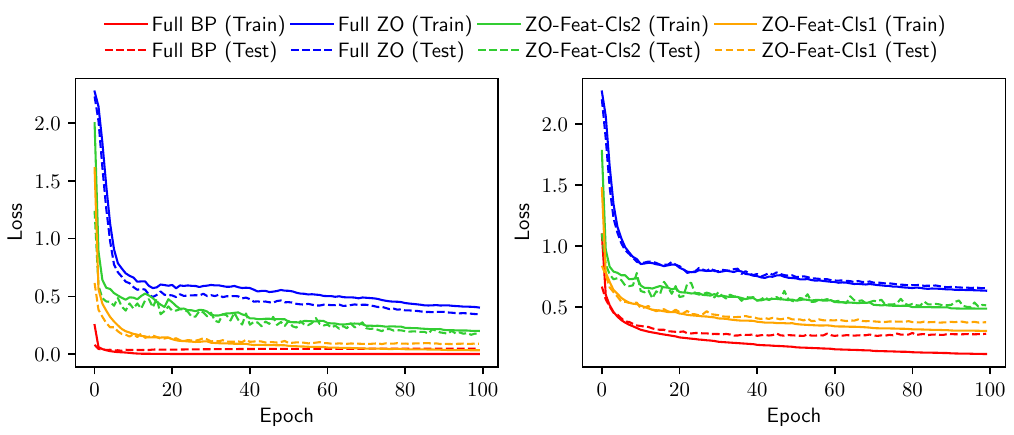}
  \caption{Training and test loss curves of LeNet-5 (FP32; left: MNIST, right: Fashion-MNIST).}
  \label{fig:loss-curve-mnist}
\end{figure}

\begin{figure}[h]
  \centering
  \includegraphics[keepaspectratio, width=0.8\linewidth]{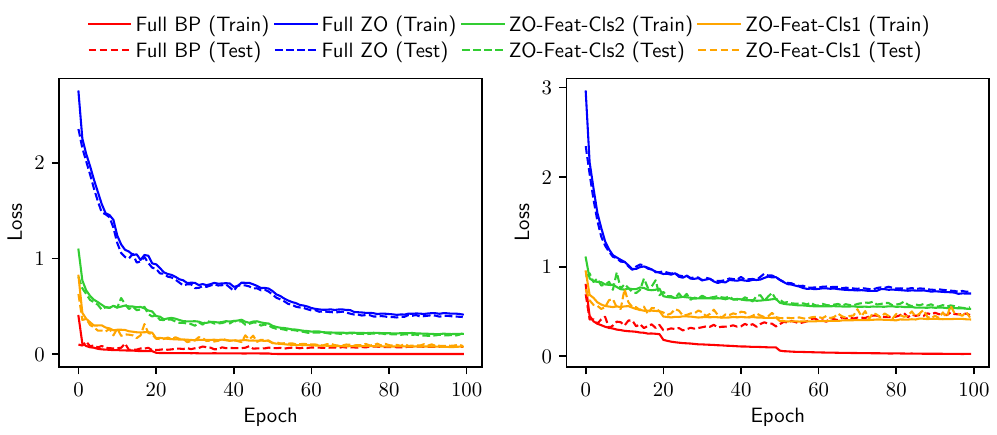}
  \caption{Training and test loss curves of LeNet-5 (INT8; left: MNIST, right: Fashion-MNIST).}
  \label{fig:loss-curve-mnist-int8}
\end{figure}

Fig. \ref{fig:loss-curve-mnist} shows training and test loss curves of the full-precision LeNet-5 on MNIST (left) and Fashion-MNIST (right).
While Full ZO suffers from a slow convergence speed and seems to require far more than 100 epochs, {\MethodName} is able to speed up the training process and close the performance gap to Full BP.
Besides, ZO-Feat-Cls1 shows less fluctuations in the loss value compared to Full ZO, indicating that training the last FC layers with BP can stabilize training.
Fig. \ref{fig:loss-curve-mnist-int8} plots loss curves of the INT8 LeNet-5 over 100 epochs.
The proposed {\MethodNameInt} successfully trains LeNet-5 and achieves a loss close to that of Full BP on both datasets.
{\MethodNameInt} has a significantly lower loss at early epochs compared to Full ZO and thus our hybrid approach is more effective in case of limited precision.
Note that the loss drops sharply at around 20 and 50 epochs, which is possibly due to changes in the hyperparameter settings ($p_\text{zero}$, Sec. \ref{sec:eval-setup-training}).

\subsection{Memory-Efficiency} \label{sec:eval-memory-eff}
\begin{figure}[h]
  \centering
  \includegraphics[keepaspectratio, width=0.9\linewidth]{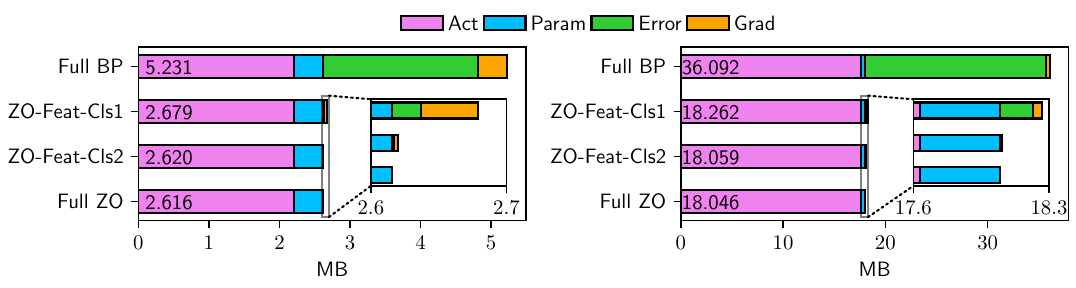}
  \caption{Memory usage breakdown of LeNet-5 (FP32; left: $B = 32$, right: $B = 256$).}
  \label{fig:memory-usage}
\end{figure}

Fig. \ref{fig:memory-usage} shows the memory usage breakdown of full-precision {\MethodName} in comparison with Full BP and Full ZO computed using Eqs. \ref{eq:memory-cost-full-bp}--\ref{eq:memory-cost-halfzo}.
We consider LeNet-5 trained on MNIST with a batch size of $B = 32, 256$.
As discussed in Sec. \ref{sec:method-memory-eff}, Full BP has the largest memory footprint, because it needs to store gradients and errors $\vb*{g}, \vb*{e}$ for all layers as well as parameters and activations $\vb*{\theta}, \vb*{a}$ (i.e., twice the memory for inference).
On the other hand, Full ZO requires half the memory of Full BP (i.e., same as inference) regardless of the batch size (5.2/2.6MB for $B = 32$, 36.1/18.0MB for $B = 256$).
ZO-Feat-Cls2 and ZO-Feat-Cls1 achieve a higher prediction accuracy close to that of Full BP (Tables \ref{tbl:accuracy}--\ref{tbl:accuracy-fine-tuning}) while consuming slightly more memory than Full ZO (+0.17/+2.4\% for $B = 32$, +0.072/+1.2\% for $B = 256$), thereby improving the accuracy-memory trade-off.
As shown in the inset figures, the additional memory usage (4.6/65.3KB for $B = 32$, 13.3/221.0KB for $B = 256$) comes from gradients and errors of the last one or two FC layers, while they account for a negligible portion.
The size of activations $\vb*{a}$ and errors $\vb*{e}$ grows linearly with the batch size $B$ and is 42.9x larger than that of parameters $\vb*{\theta}$ when $B = 256$.
This highlights the memory-efficiency of ZO methods that do not need to save large error tensors for BP.
The activations still dominate the memory consumption and take up 97.6\% and 96.6\% in ZO-Feat-Cls2 and ZO-Feat-Cls1 with $B = 256$.

\begin{figure}[h]
  \centering
  \includegraphics[keepaspectratio, width=0.9\linewidth]{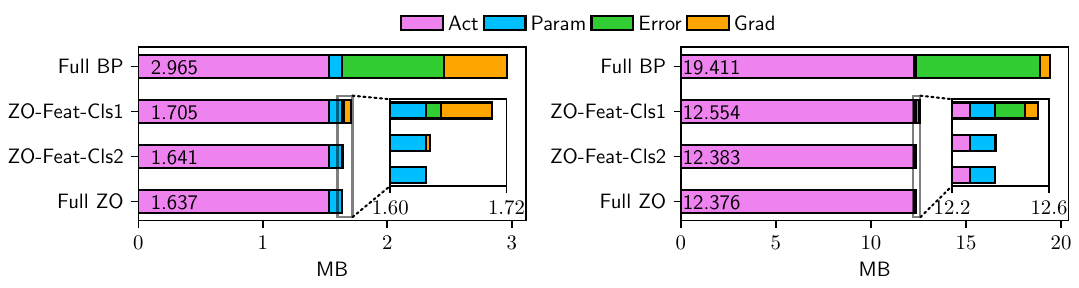}
  \caption{Memory usage breakdown of LeNet-5 (INT8; left: $B = 32$, right: $B = 256$).}
  \label{fig:memory-usage-int8}
\end{figure}

Fig. \ref{fig:memory-usage-int8} shows the memory cost of {\MethodNameInt} as well as that of Full BP and Full ZO.
In this case, we use Eqs. \ref{eq:memory-cost-full-bp-int8}--\ref{eq:memory-cost-halfzo-int8} to compute the memory cost required to train INT8 LeNet-5 on MNIST ($B = 32, 256$).
Similar to the FP32 case (Fig. \ref{fig:memory-usage}), Full ZO consumes the least amount of memory (1.8/1.6x less than Full BP for $B = 32, 256$) but has the lowest prediction accuracy (Table \ref{tbl:accuracy}).
On the other hand, ZO-Feat-Cls2 and ZO-Feat-Cls1 achieve a comparable accuracy to Full BP while maintaining nearly the same memory cost as Full ZO (0.052/1.4\% and 0.26/4.1\% more than Full ZO with $B = 32, 256$).
As discussed in Sec. \ref{sec:method-int8-memory-eff}, FC and convolution layers need extra buffers to store intermediate accumulation results in 32-bit integers $\vb*{a}_l^\text{int32}, \vb*{g}_l^\text{int32}, \vb*{e}_{l - 1}^\text{int32}$ when computing activations, gradients, and errors.
While the size of gradients and errors is substantially reduced in {\MethodNameInt} and Full ZO, activations account for the largest portion of the entire memory usage (93.5\% and 90.0\% in ZO-Feat-Cls2 and ZO-Feat-Cls1 with $B = 32$).
Compared to the FP32 counterparts (Fig. \ref{fig:memory-usage}), INT8 ZO methods (ZO-Feat-Cls1, ZO-Feat-Cls2, Full ZO) require 1.46--1.60x less memory, which is below an ideal case (4x) due to extra buffers $\vb*{a}_l^\text{int32}, \vb*{g}_l^\text{int32}, \vb*{e}_{l - 1}^\text{int32}$.
While we assume that all necessary variables are allocated at once and not released during training, they can be dynamically allocated and deallocated as necessary to further optimize the memory usage, which is a future work.

\begin{figure}[h]
  \centering
  \includegraphics[keepaspectratio, width=0.5\linewidth]{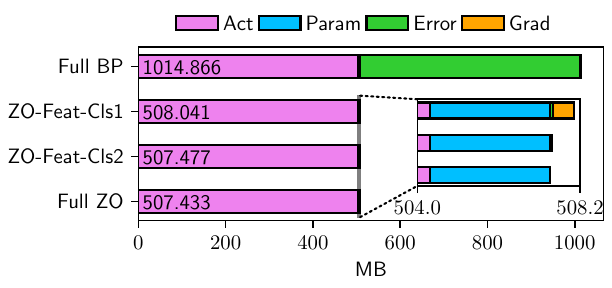}
  \caption{Memory usage breakdown of PointNet (FP32, $B = 32$).}
  \label{fig:memory-usage-pointnet}
\end{figure}

Fig. \ref{fig:memory-usage-pointnet} shows the memory usage when training PointNet on ModelNet40.
The size of activations and errors is significantly larger than that of LeNet-5 (Fig. \ref{fig:memory-usage}) and dominates the overall memory cost (99.4\% and 99.3\% in ZO-Feat-Cls2 and ZO-Feat-Cls1).
As depicted in Fig. \ref{fig:half-zo}, PointNet extracts a high-dimensional local feature for each point using a stack of five FC layers, and then aggregates these per-point features into a global feature via max-pooling.
The last FC layer in the feature extraction part produces an activation of size $(B, N, 1024)$, with $B, N$ being the batch size and number of points ($N = 1024$ in ModelNet40).
While {\MethodName} needs to keep gradients and errors of the last FC layers of the classification head, they are barely noticeable in Fig. \ref{fig:memory-usage-pointnet} and do not significantly affect memory requirements (0.0087\% and 0.12\% for ZO-Feat-Cls2 and ZO-Feat-Cls1).
As a result, {\MethodName} almost halves the memory cost while achieving a comparable accuracy to Full BP (i.e., vanilla SGD) in the point cloud classification task as well.

\subsection{Execution Time} \label{sec:eval-exec-time}
\begin{figure}[h]
  \centering
  \includegraphics[keepaspectratio, width=0.95\linewidth]{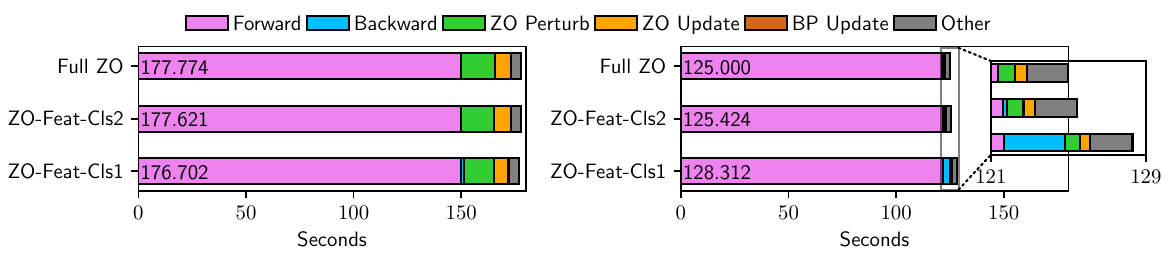}
  \caption{Execution time breakdown of the C++ implementation of Full ZO and {\MethodName} on Raspberry Pi Zero 2 (left: FP32, right: INT8).}
  \label{fig:execution-time}
\end{figure}

We evaluate the execution time of {\MethodName} and {\MethodNameInt} in comparison with Full ZO.
Fig. \ref{fig:execution-time} shows the results obtained on a Raspberry Pi Zero 2 board using our C++ implementation.
We train the FP32 and INT8 LeNet-5 on MNIST for 100 epochs and measure the average execution time per epoch.
Note that the breakdown follows the algorithms presented in Algs. \ref{alg:halfzo}--\ref{alg:halfzo-int8} (e.g., \textbf{Forward} and \textbf{ZO Update} correspond to lines \ref{alg:halfzo-forward0}+\ref{alg:halfzo-forward1} and \ref{alg:halfzo-perturb0}+\ref{alg:halfzo-perturb1} in Alg. \ref{alg:halfzo}).
Importantly, both {\MethodName} and {\MethodNameInt} maintain the training speed of Full ZO while significantly improving the accuracy.
As expected, forward passes account for the largest portion of the wall-clock execution time and determine the overall performance (84.3--84.8\% and 94.8--97.1\% for {\MethodName} and {\MethodNameInt}), whereas backward passes are almost negligible ($<$0.76\% and $<$2.5\%).
This indicates that existing inference engines can be used to boost the performance of training with {\MethodName} and {\MethodNameInt}.
Besides, {\MethodNameInt} is 1.38--1.42x faster than {\MethodName} thanks to the use of low-precision arithmetic (FP32 SIMD instructions can be replaced by INT8 instructions that offer more parallelism).
The parameter perturbation and update in ZO (\textbf{ZO Perturb} and \textbf{ZO Update}) are inherently memory-intensive and account for a non-negligible portion of the execution time (11.8--13.1\%) in the FP32 case (left).
On the other hand, they only account for 1.0--1.2\% in the INT8 case (right), which can be attributed to 4x reduction of the parameter size and improved cache efficiency.

\section{Conclusion} \label{sec:conc}
In this paper, we proposed {\MethodName} and {\MethodNameInt} as lightweight on-device learning methods that leverage memory-efficiency of ZO-based training and high accuracy of BP-based training.
The key idea of {\MethodName} is to employ BP for the last few layers and ZO for the remaining part.
We devised a method to estimate quantized ZO gradients based on the integer cross-entropy loss, thereby allowing {\MethodNameInt} to perform ZO-based training using only 8-bit integer arithmetic.
We formulated the memory usage of {\MethodName} and {\MethodNameInt} to analyze their memory-efficiency.
The evaluation on image and point cloud classification datasets demonstrated that, both {\MethodName} and {\MethodNameInt} achieve comparable prediction accuracy to Full BP by training only the last few FC layers with BP instead of ZO.
They consumed nearly the same amount of memory and maintained the execution time as ZO methods, thereby significantly improving the trade-off between accuracy and training cost.
On LeNet-5 and Fashion-MNIST, {\MethodName} and {\MethodNameInt} provided 5.2--9.5\% and 6.4--10.7\% higher accuracy than the vanilla ZO with a 0.072--0.17\% and 0.052--0.26\% memory overhead.
{\MethodNameInt} reduced the memory cost by 1.46--1.60x and execution time by 1.38--1.42x compared to {\MethodName} without compromising the accuracy.
Besides, {\MethodName} and {\MethodNameInt} are successfully applied to fine-tuning as well as training from scratch.

While being successful on simple DNN architectures and classification datasets, there is still an accuracy gap between {\MethodName} and BP-based training that need to be addressed in a future work.
In addition, it is an open problem to determine the number of layers trained by BP such that {\MethodName} strikes the balance between accuracy and training cost.
We also plan to apply {\MethodName} to large-scale DNNs including language models and image diffusion models to evaluate the practicality of our approach.
It is also possible to combine {\MethodName} with PEFT techniques such as LoRA~\cite{EdwardHu22} and QLoRA~\cite{TimDettmers23}.


\renewcommand{\baselinestretch}{1.0}
\bibliographystyle{unsrt}


\vfill

\end{document}